\documentclass{article}
\pdfoutput=1

\usepackage[utf8]{inputenc}
\usepackage{makecell}
\usepackage{subcaption}
\usepackage{booktabs}
\usepackage{multirow,bigdelim}
\usepackage{makecell}
\usepackage[verbose]{placeins}
\usepackage{amsmath}
\usepackage[most]{tcolorbox}
\usepackage{siunitx}

\usepackage{graphicx}

\usepackage{pgfplots}
\usepgfplotslibrary{groupplots}
\usetikzlibrary{patterns,shapes.arrows, positioning, matrix}
\pgfplotsset{compat=newest}
\DeclareUnicodeCharacter{2212}{−}

\newcommand{\ra}[1]{\renewcommand{\arraystretch}{#1}}

\title{An Evaluation of DNN Architectures for Page Segmentation of Historical Newspapers}

\author{
  Bernhard Liebl \\
  \texttt{Computational Humanities Group, Leipzig University}\\
  \texttt{liebl@informatik.uni-leipzig.de}
  \and
  Manuel Burghardt \\
  \texttt{Computational Humanities Group, Leipzig University}\\
  \texttt{burghardt@informatik.uni-leipzig.de}
}
\date{\today}

\begin{document}

\maketitle

\begin{abstract}
One important and particularly challenging step in the  optical character recognition (OCR) of historical documents with complex layouts, such as newspapers, is the separation of text from non-text content (e.g. page borders or illustrations). This step is commonly referred to as page segmentation. While various rule-based algorithms have been proposed, the applicability of Deep Neural Networks (DNNs) for this task recently has gained a lot of attention. In this paper, we perform a systematic evaluation of 11 different published DNN backbone architectures and 9 different tiling and scaling configurations for separating text, tables or table column lines. We also show the influence of the number of labels and the number of training pages on the segmentation quality, which we measure using the Matthews Correlation Coefficient. Our results show that (depending on the task) Inception-ResNet-v2 and EfficientNet backbones work best, vertical tiling is generally preferable to other tiling approaches, and training data that comprises 30 to 40 pages will be sufficient most of the time.
\end{abstract}

\section{Introduction}
With the emergence of the digital humanities, historical documents and especially newspapers and magazines are becoming increasingly popular for quantitative analyses by means of text mining and distant reading \cite{graham2015exploring}.
Accordingly, the digitization of these historical documents poses very specific challenges in the area of layout analysis, since the documents often have a complex and very heterogeneous structure. Furthermore, the quality of historical sources is typically far from optimal.

In this paper we describe insights from a current project that is concerned with the digitization and analysis of one of those historical newspapers, the \textit{Berliner Börsen-Zeitung} (BBZ)\footnote{The Berliner Börsen-Zeitung was a German newspaper published between 1855 and 1944.}. 
As with any text document, various optical character recognition procedures need to be applied in order to digitize the newspaper.
At its most basic level, this involves the conversion of images of single characters (or lines of characters) into some form of encoded text. 
To do this, however, the text parts of a document must first be extracted. This process is called page segmentation and can be understood as part of a larger task called layout detection.

Since any subsequent OCR computation depends on the quality of this data, page segmentation is a crucial factor in the overall quality of an OCR pipeline. While there are some proved and tested approaches for page segmentation of text-only documents, the problem becomes considerably more difficult with so called mixed-content documents that contain text as well as images and other non-text content \cite{shafait_dissertation, Winder2010ExtendingTP}.

The BBZ, on the other hand, not only contains texts, images, separator lines, and embellishing borders, but also various complex, multi-column table structures. We want to differentiate all of these elements and separate them from the actual text at an early layout processing stage.
For this purpose, we present an evaluation of Deep Neural Network (DNN) architectures that can perform pixel-wise semantic segmentation for the BBZ. Since the BBZ exhibits many typical challenges and problems of historical newspapers, including complex layouts, paper degradation as well as warping and tearing of pages, we believe the results of our evaluation will be relevant for page segmentation tasks on other historical newspapers as well.

\section{Related Work}
In addition to several non-neural procedures on the semantic segmentation of images (see e.g. \cite{Winder2010ExtendingTP, thoma_survey_2016, shafait_performance_2008}), more recently DNNs have established themselves as another approach to this task.
In the context of OCR, pixel-by-pixel segmentation using DNNs has been applied successfully to historical documents \cite{dhsegment_2018, wick_fully_2018, chen_convolutional_2017} as well as to newspapers \cite{meier_fully_2017}. Two of the most commonly used libraries for this task, dhSegment \cite{dhsegment_2018} and P2PaLA, \cite{p2pala2017} also make use of DNN architectures (ResNet and GANs). However, one problem with pixel-by-pixel segmentation is the accidental merging of adjacent regions. Alternative network architectures like Mask R-CNNs \cite{he_mask_2018} might be able to alleviate this issue.

In the field of pixel-by-pixel segmentation, the number of new promising standard DNN architectures built for segmentation and/or visual recognition tasks has seen a considerable increase recently \cite{dhsegment_2018, liu_recent_2019, khan2019survey}. Except for dhSegment though, most previous research on using DNNs for historical document analysis seems to have been focused on designing custom network architectures and training networks from scratch \cite{wick_fully_2018, chen_convolutional_2017, meier_fully_2017}. Following the approach of dhSegment, we apply transfer learning \cite{yosinski_how_nodate} to architectures that have been pretrained on ImageNet \cite{russakovsky_imagenet_2015}. Transfer learning has been repeatedly shown to "improve generalization performance even after substantial fine-tuning on a new task" \cite{yosinski_how_nodate}, which is highly desirable for historical document analysis. Unlike dhSegment though, we do not focus on a single architecture (ResNet-50) with a static configuration. 

On the whole, there seems to be a lack of research that compares and evaluates different pixel segmentation architectures, especially when used with transfer learning on historical documents. While single components of the network, such as the quality of the so-called backbone networks, are well understood \cite{russakovsky_imagenet_2015, bianco_benchmark_2018}, the influence of other parameters such as the number of labels or the handling of scaling and tiling, are lesser researched. The latter is particularly important in the context of historical documents though, since image resolutions tend to be rather high, and some form of scaling or tiling is necessary with current GPU hardware. Still, most researchers pick these parameters rather arbitrarily, without having data for supporting one or the other decision. For example, the paper on dhSegment \cite{dhsegment_2018} investigates two specific tile sizes, $400 \times 400$ and $600 \times 600$ for two different documents, but does not compare the quality of the two. 

In short: there have been numerous successful applications of specific DNN architectures trained from scratch for historical document segmentation. Yet there seems to be only little research (most notably dhSegment) on the evaluation of transfer learning on commonly used, pre-trained architectures, especially in conjunction with an investigation of the important effects of tiling, scaling and the number of labels in the context of historical documents. 
In this paper, we try to find answers to these open questions by providing a systematic evaluation of DNN architectures for the page segmentation of historical newspapers.

\FloatBarrier

\section{Segmentation Tasks}

Next we describe the actual segmentation tasks we aim to accomplish. We also provide an overview of how we created a ground truth for the specified tasks.

\subsection{Specification of Tasks}

For our evaluation study, we split the overall segmentation task into three smaller tasks named \emph{blk}, \emph{blkx} and \emph{sep}. These tasks are capable of predicting semantic classes like text or tables (see Table \ref{table:segment_classes} and Figure \ref{figure:pagegroundtruth}): \emph{blk} separates background, text regions and table regions, \emph{blkx} is an extension of \emph{blk} that also recognizes images and borders\footnote{We split \emph{blkx} from \emph{blk} to perform an ablation study on how much more difficult the overall segmentation task gets when differentiating between empty background and non-text background content such as decorated borders and images.}. The task \emph{sep} detects straight or slightly curved separator lines between text regions and inside tables that are already present in the layout. These might be useful for layout detection engines later.

\FloatBarrier

\noindent\begin{minipage}{\linewidth}
\centering

\vspace{0.5cm}

    \centering \scriptsize
	\ra{1.0}
\begin{tabular}{@{}llllll@{}}\toprule

	class & meaning & blk & blkx & sep & \\ 
\midrule

BACKGROUND & background (no other label) & x & x & x\\
TXT & regions of text context & x & x & \\
TAB & regions of table content & x & x &  \\
ILLUSTRATION & regions with pictures or borders & & x & \\
H & horizontal separators & & & x \\
V & vertical separators & & & x  \\
T & vertical column separators in tables & & & x \\

	\end{tabular}
    \captionof{table}{\label{table:segment_classes}Segmentation tasks and label classes.}	
    
\vspace{0.5cm}

    \fbox{
		\includegraphics[scale=0.12]{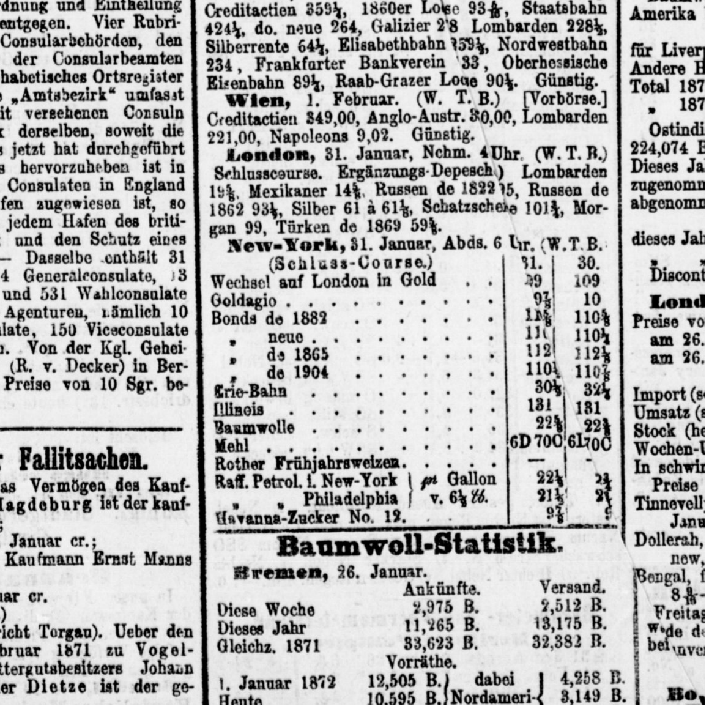}
    }   
    \hspace{10pt}
    \fbox{
		\includegraphics[scale=0.12]{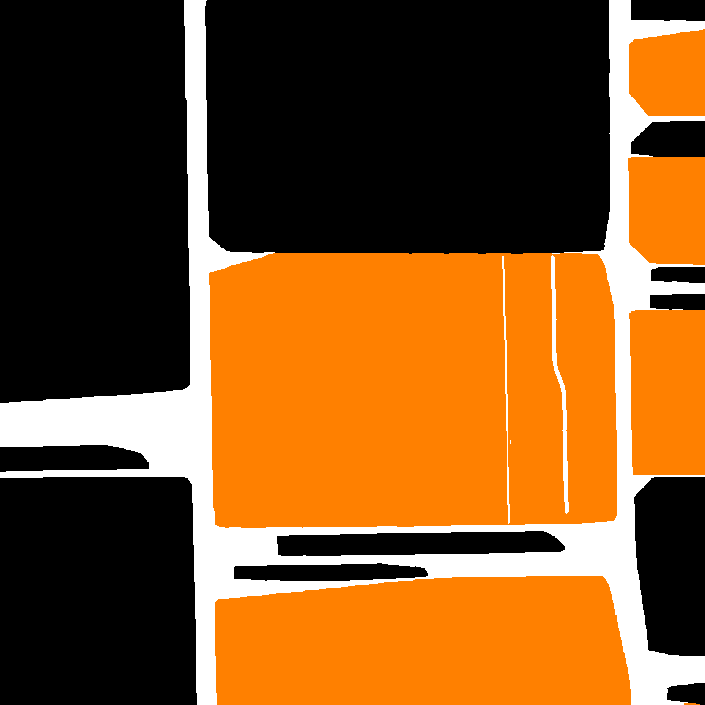}
    }
    \hspace{10pt}
    \fbox{
		\includegraphics[scale=0.12]{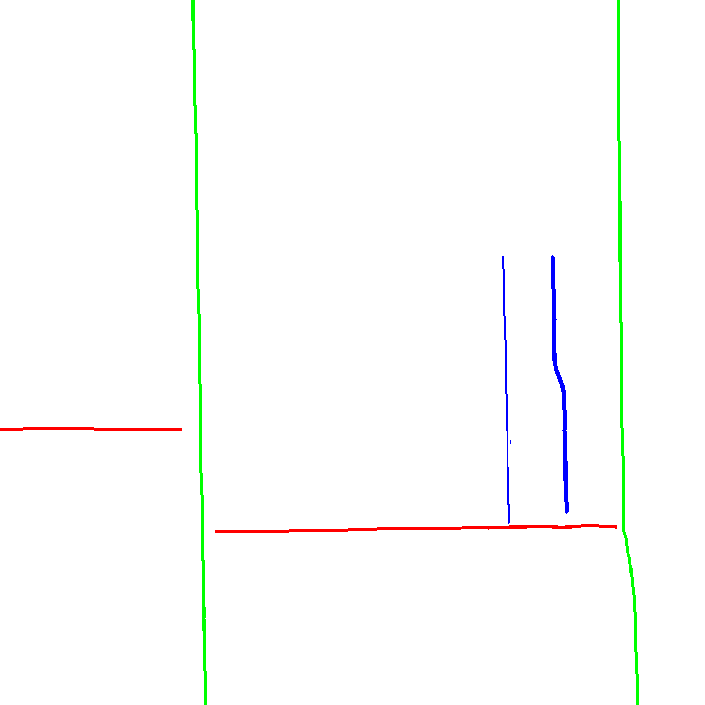}
    }

    \captionof{figure}{Left: Small section of a BBZ page from February 1, 1872. Middle: Ground truth for the same page for task \emph{blk}, with 3 classes \emph{BACKGROUND} (white), \emph{TXT} (black) and \emph{TAB} (orange). Right: Ground truth for the same page for task \emph{sep}, with 4 classes \emph{BACKGROUND} (white), \emph{H} (red), \emph{V} (green), and \emph{T} (blue).}

	\label{figure:pagegroundtruth}

\end{minipage} 

\FloatBarrier

\subsection{Manual Generation of Ground Truth}

\begin{figure}
    \fbox{
		\includegraphics[scale=0.12]{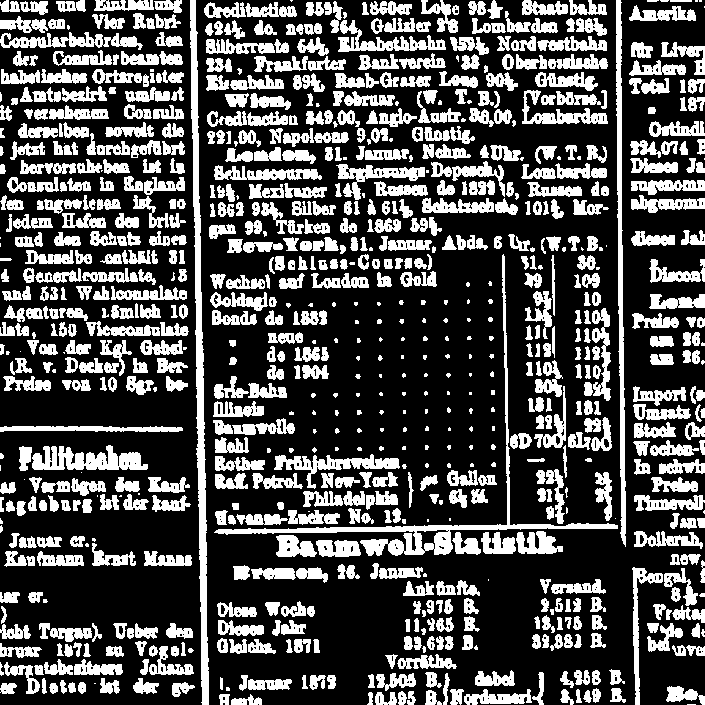}
    }   
    \hspace{10pt}
    \fbox{
		\includegraphics[scale=0.12]{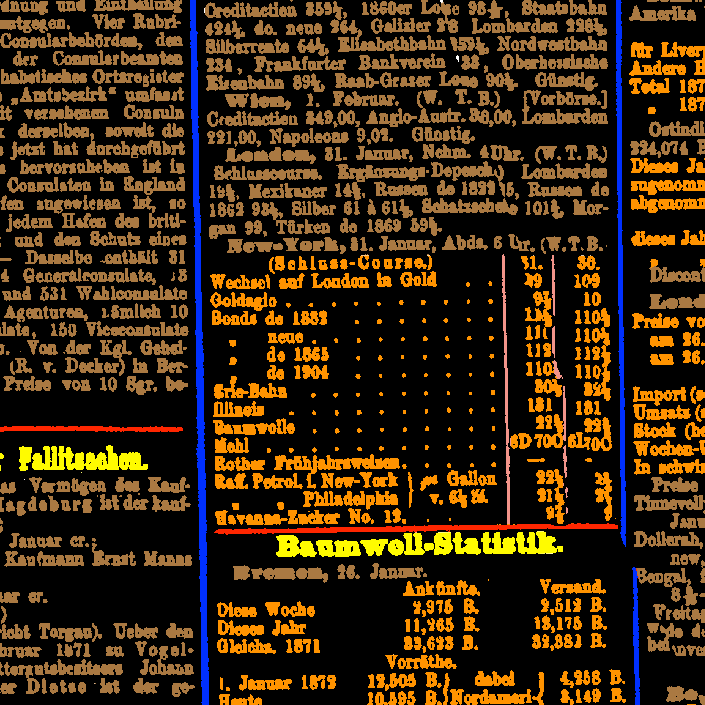}
	}
    \hspace{10pt}
    \fbox{
		\includegraphics[scale=0.17]{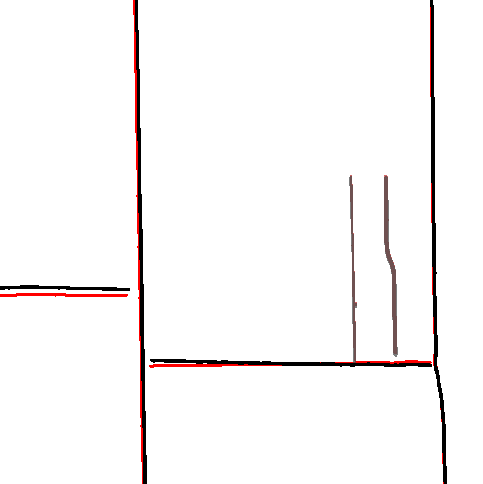}
    }
	\caption{
		Left: Binarized version of BBZ page. Middle: Manually annotated version before preprocessing (using various label classes that were later combined to the ones given in the previous section). Right: Separator ground truth for warped (black) and unwarped (red) version of the image.
	}
	\label{figure:ann}
\end{figure}

Our corpus contains a total of 642,480 scanned pages from the aforementioned BBZ. For generating ground truth data, we sampled 104 pages, which were selected randomly using a script.
First, pages were binarized using ocropus \emph{nlbin} \cite{iwamura_robust_2014} with custom settings\footnote{Settings: threshold=0.6, escale=5,
            border=0.1,
            perc=80,
            range=20,
            low=20 and 
            high=80.}. 
Next, annotations were added in full resolution onto the binarized images by manually applying colored drawings onto a layer with its blending mode set to \emph{multiply} in a standard photo editing application (see Figure \ref{figure:ann}). Different labels were annotated with different colors and each pixel received a unique label. Annotations that ended up on binarized black regions were ignored in succeeding steps. Depending on the page complexity, this step took about one to four hours per page.

The resulting RGB images were then converted into indexed color images. An improved workflow might restrict annotation images to indexed colors from the beginning. We implemented some basic checks that prevent unintended RGB colors to enter the image.
            
\subsection{Postprocessing Ground Truth}

The better the quality of the ground truth, the better a DNN can learn to deal with ambiguous situations. For this reason, we performed a number of rule-based normalization procedures on our manual annotations that aimed at generalizing them and removing noise that is inherent in manual, pixel-wise annotation. More concretely, we used rule-based cleanups tuned for our specific corpus\footnote{Details of this can be found in our implementation, available at https://github.com/poke1024/bbz-segment.}. 
For example, for text and table blocks, we morphologically closed the annotated blocks into solid shapes without connecting adjacent blocks. For separators we detected broken lines and reconnected them, up to a certain threshold. We used heuristic rules to decide whether two blocks which got merged through a morphological closing operation should actually stay merged, based on their respective line heights (the latter were computed by running Tesseract on the polygonal text block's image). We manually verified all these heuristic operations on our training data.

In the next step, images were scaled down to their intended image size (defined later below as part of our tiling configurations, see Table \ref{table:tilings}) during training. Since scaling schemes like linear interpolation or nearest neighbor would destroy indexed label information,  our scaling approach used an area filter that gave weights to labels and then picked the one label in each source pixel area that had the highest summed weight. By choosing higher weights for separator labels, we ensured that their thin line annotations did not get destroyed or broken during the scaling process.

For data augmentation, we also added slightly warped versions of each image. We modified source images and annotations by identical projections onto a mesh derived from pseudorandom cubic splines. Once again, this operation used special area filters for scaling. We basically reversed the approach described in \cite{zucker} (see Figure \ref{figure:ann} for an illustration). Note that this  data augmentation was independent of the later data augmentation during the training.

Both warped and non-warped images were then scaled and/or split into overlapping tiles, as described in section \ref{Tiling}. Since folds were defined over pages (not tiles), so all tiles belonged to either validation or training set, i.e. there was no danger of information leakage between training and validation sets.

\section{Training the DNN Model}

In this section, we provide a review of existing DNN architectures and explain our choice of an architecture that -- given the accompanying ground truth -- is capable of learning and predicting our previously described segmentation tasks with the best possible quality. 

\subsection{Architectures and Hyperparameters}

There are many parameters to consider when training a DNN for the specified tasks. In the following sections, we consider some fundamental options and explain how we came up with an optimal configuration that best suits our needs.

\subsubsection{Configuration Space}

Semantic segmentation networks consist of two architectures: (1) a \emph{backbone} network that performs the visual recognition tasks and that acts as an encoder, and (2) an overall architecture for the segmentation network as a whole \cite{liu_recent_2019}, to which we will refer to as \emph{s-model} (for segmentation model) from here on. For example, ResNet \cite{he_deep_2015} is a well-established recognition \emph{backbone} used in many real-world U-Net implementations, whereas U-Net \cite{ronneberger-u-net} is a well-established \emph{s-model} architecture.

In our search for a specific segmentation DNN (i.e. a network consisting of both a \emph{backbone} and a \emph{s-model}), and following the terminology of configuration spaces as described by Feurer and Hutter \cite{feurer-automlbook19a}, we defined a configuration space ${\mathbf{\Theta}}$ as \begin{equation} \mathbf{\Theta}=\Theta_{s-model} \times \Theta_{backbone} \times \Theta_{tiling} \times \Theta_{optimizer} \times \Theta_{loss}  \end{equation}

$\Theta_{optimizer}$ is the optimization algorithm used for training the model, $\Theta_{loss}$ is the DNN's loss function during training.

\subsubsection{Evaluation Metric}

There are various measures at hand when it comes to evaluating the pixel-segmentation tasks, among them the well-known F1 score \cite{Yakubovskiy}, while others are based on pixel accuracy and intersection over union (IOU) \cite{long_fully_2014}. However, it is unclear which of these measures is best suited as an indicator for overall quality in a ranking.\footnote{See Table \ref{table:evaluation} for an illustration of this. This has also been noted before: "[o]ne of the main issues when comparing different neural networks architectures is how to select an appropriate metric to evaluate their accuracy", as "commonly employed evaluation metrics can display divergent outcomes, and thus it is not clear how to rank different image segmentation solutions." \cite{fernandez-moral_new_2018}}.
In the context of OCR of historical documents, Wick and Puppe \cite{wick_fully_2018} report issues with Long et al.'s 
proposed metrics and suggest an alternative pixel accuracy metric. At the same time, it is known that pixel accuracy is generally not well suited for imbalanced label classes like our \emph{blkx} and \emph{sep} tasks.\footnote{Chicco and Jurman note: "when the dataset is unbalanced (the number of samples in one class is much larger than the number of samples in the other classes), accuracy cannot be considered a reliable measure anymore" \cite{chicco_advantages_2020}.} 

From this situation, we decided to evaluate the overall quality of our models using an altogether different metric, namely a multi-class extension of the Matthews Correlation Coefficient (MCC). For binary classification Chicco and Jurman report that "F1 score and accuracy can mislead, but MCC does not" \cite{chicco_advantages_2020}, and note that, in a ranking experiment of various classifier algorithms "[t]he F1 score ranking and the accuracy ranking, in conclusion, are hiding [an] important flaw of the top classifier" \cite{chicco_advantages_2020}.\footnote{Specifically, the reported flaw was a sub-excellent true negative rate in the top ranked classifier.}
While we are aware that for pixel segmentation, MCC does not seem to be a widely used metric at the moment, and that other metrics have also been studied and proposed in this context \cite{thoma_survey_2016, fernandez-moral_new_2018, csurka_what_2013}, we think the arguments in \cite{chicco_advantages_2020} are very plausible and  believe that the MCC is also well-suited for our ranking experiments on highly unbalanced label classes.

\subsubsection{Grid Search: General Approach}

In order to identify the optimal configuration of hyperparameters, we performed a grid search. For both the grid search as well as the final model we made use of models pretrained on ImageNet \cite{russakovsky_imagenet_2015, Yakubovskiy}.
All models were trained on our full training data for 50 epochs with a batchsize of 3 and a learning rate of $\num{2.5e-3}$. For evaluation we only used fold 1 of an 80-20 split between training data and validation data. Furthermore, we used automatic data augmentation via \cite{2018arXiv180906839B} with (1) rotation of up to $\pm10$ degrees (applied always) and (2) contrast manipulation with a probability of 90\% using default settings.

\subsubsection{Choices for $\Theta_{optimizer}$ and $\Theta_{loss}$}

In a first step, we pinned $\Theta_{s-model}$ to ResNet-18, $\Theta_{backbone}$ to U-Net, and $\Theta_{tiling}$ to an untiled 896 x 128 scaling, to find good values for $\Theta_{optimizer}$ and $\Theta_{loss}$.
For $\Theta_{loss}$, categorical cross-entropy loss proved best for \emph{blk} and \emph{blkx}, whereas generalized dice loss \cite{sudre_generalised_2017} showed the best results for \emph{sep}.
For $\Theta_{optimizer}$ we obtained very stable convergence and good results with an Adam optimizer with decoupled weight decay regularization \cite{loshchilov_decoupled_2017}, lookahead \cite{zhang_lookahead_2019} and cosine annealing. Various other current optimizers we tried showed numerical instabilities with our training data.

In a second step, we pinned $\Theta_{optimizer}$ and $\Theta_{loss}$ and performed a grid search over $\Theta_{s-model}$, $\Theta_{backbone}$ and $\Theta_{tiling}$ (see the following sections).

\subsubsection{Choices for $\Theta_{s-model}$}

$\Theta_{s-model}$ included U-Net \cite{ronneberger-u-net} and FPNs \cite{liu_recent_2019}. We also evaluated LinkNet \cite{chaurasia_linknet_2017} and PSPNet in a large prior grid search, but they gave consistently worse results than U-Net and FPN, so for the sake of minimizing the already huge grid evaluation runtimes, we decided to exclude those models from the actual evaluation. 

\subsubsection{Choices for $\Theta_{backbone}$}

Based on \cite{bianco_benchmark_2018}, our batch size, and GPU memory constraints, we picked the following architectures as representative of all possible choices: VGG16, ResNet-34, ResNet-50, SE-ResNet-34, SE-ResNet-50, SE-ResNeXt-50, Inception-v3, Inception-ResNet-v2.\footnote{We included ResNet-34 variants, since they are so popular and as such are an important performance reference.}
We also included variants of the rather new EfficientNet \cite{tan_efficientnet_2019}, namely EfficientNet-B1 and EfficientNet-B2, but omitted EfficientNet-B3 and higher since we ran out of GPU memory on our V100 with our standard batch size for them.

Note that in general we restricted ourselves to models that allowed us to carry out at least batch sizes of 3 (due to batch norm and other considerations) on all tile configurations.

\subsubsection{Choices for $\Theta_{tiling}$}
\label{Tiling}

As document images are generally large (e.g. at least $2,400 \times 3,400$ pixels in a medium-sized version of the BBZ) it is not possible to train a model on the entire, unscaled image with current GPUs. 
One important aspect thus is the scaling size of the images for training and prediction. Accordingly, three questions arise: should we (a) just scale down images (thereby losing information), (b) split images into smaller parts, i.e. into tiles or patches (thereby losing context), or (c) do both.

Unfortunately, in this general form, these three questions are rather intractable. Therefore, we need to generate specific examples that we can test and interpret. We also need to pick specific image resolutions, specific tile patterns (e.g. horizontal vs. vertical), specific GPU hardware and, finally, specific combinations of those factors that will allow us extrapolate answers to the above questions. 
In other words, we sample a small subset from a very large population of possible combinations and use these as parameters for $\Theta_{tiling}$ in our grid search.

In this manner, we first need to pick specific image resolutions to investigate. To this end, we explored the maximum (technically) possible size of an image or tile on certain GPUs (see below). Based on this maximum size, we then carefully designed a number of tiling configurations that include mixed and border cases that model important features from (a), (b) and (c).

As GPU hardware for all further tests, we picked two GPUs that seem representative of current top-tier and enterprise-grade GPU hardware and therefore are of large practical interest for institutions having one or the other: a NVIDIA GeForce RTX 2080 Ti with 12 GB GPU RAM and a NVIDIA Tesla V100 with 32 GB GPU RAM.
We explored RAM limits of those GPUs by empirically searching for resolutions that (i) matched our corpus document image aspect ratio and other side constraints\footnote{The aspect ratio of our newspaper corpus image data is about 1.45. An additional constraint is that width and height need to be multiples of 64 on many DNN platforms.}, (ii) were as large as possible, (iii) would allow us to train all models investigated in our grid search using \cite{Yakubovskiy} and (iv) use batch size 3. The largest feasible resolutions we found were $512 \times 768$ (0.39 megapixels) on the GeForce RTX 2080 Ti, and $896 \times 1280$ (1.15 megapixels) on the V100. From here on, we call these two pixel counts, i.e. 1.15M and 0.39M, \emph{maximum sizes} - as they are the \emph{maximum} number of pixels trainable on those GPUs. 
As these are the limits of current GPU hardware, they are of great practical importance and will become pivotal in the following discussion.

\begin{figure}[h]
    \fbox{
		\includegraphics[scale=0.2]{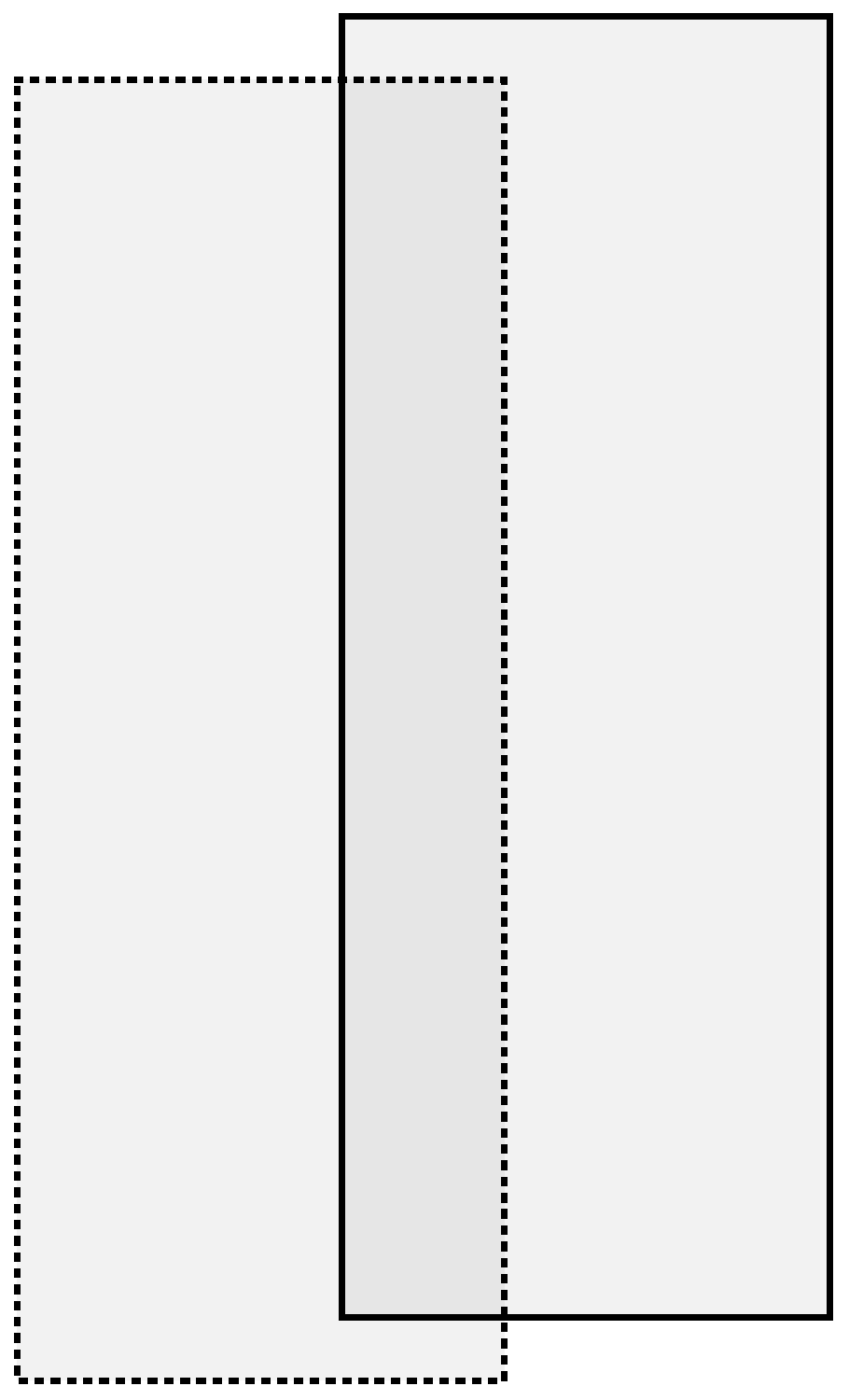}
    }   
    \hspace{50pt}
    \fbox{
		\includegraphics[scale=0.2]{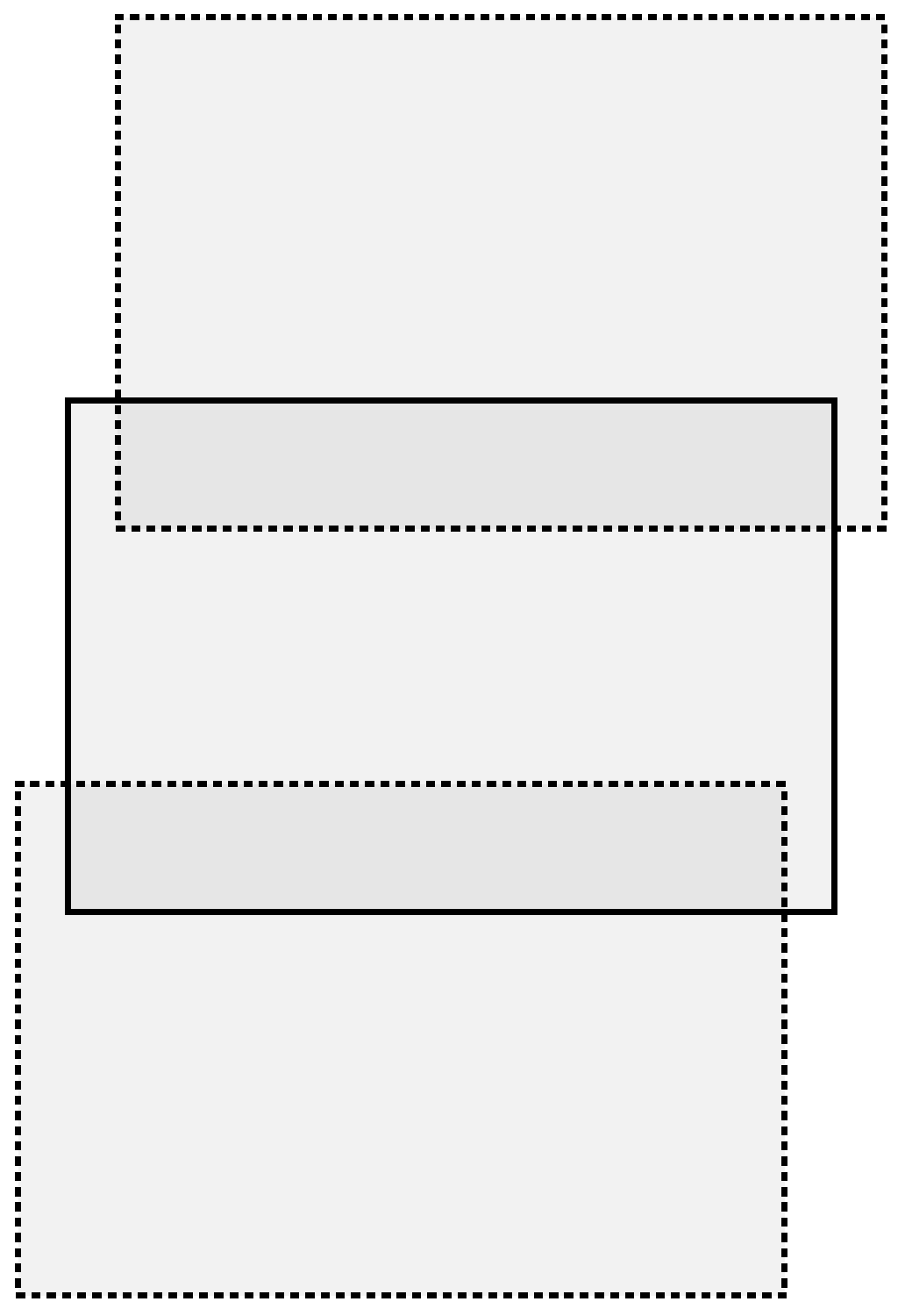}
    }
    \hspace{50pt}
    \fbox{
		\includegraphics[scale=0.2]{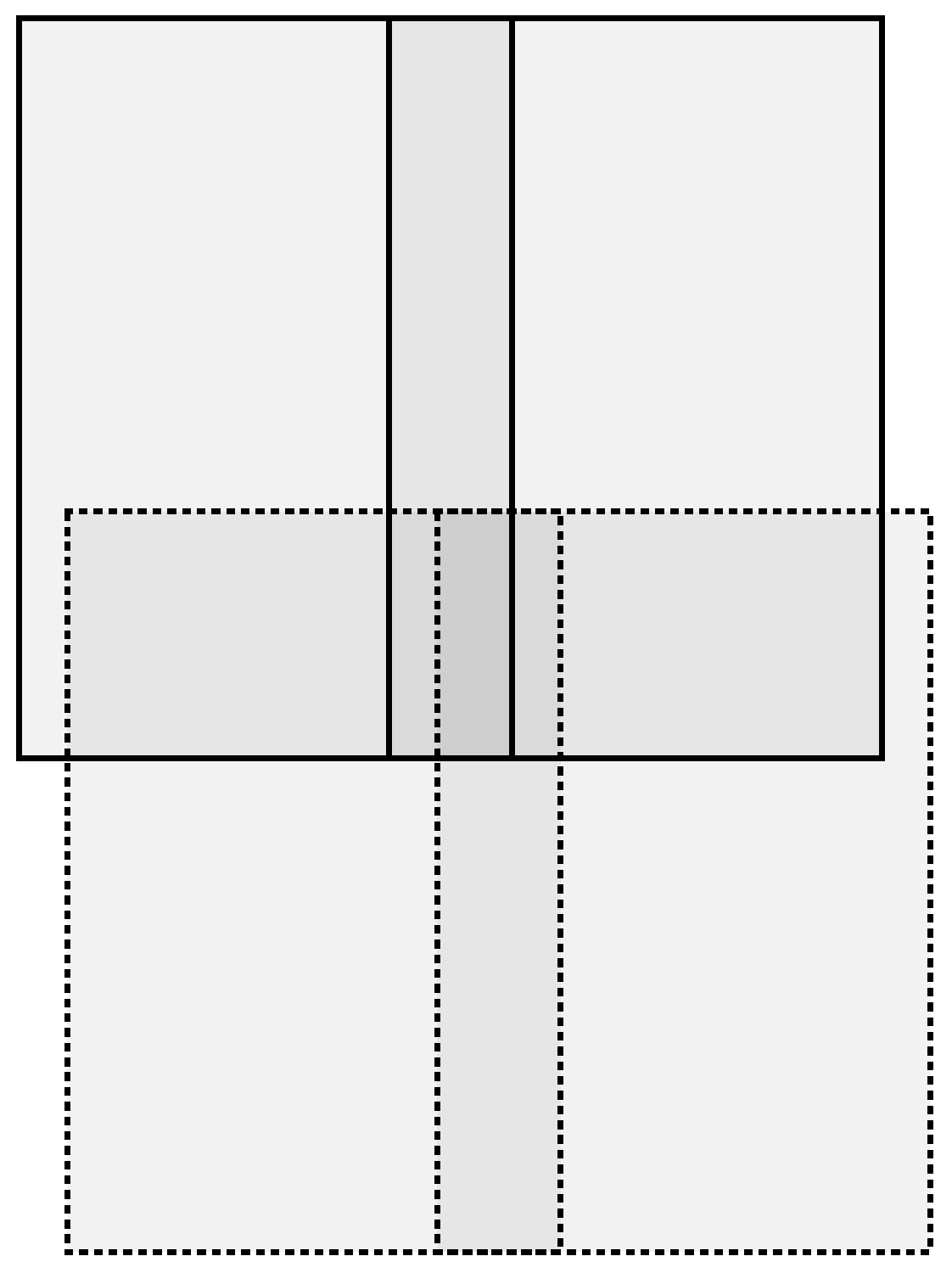}
    }
	\caption{
		Tiling patterns. Left: Horizontal tiling (h). Middle: Vertical tiling (v). Right: Rectangular tiling (h, v). Note that tiles have been slightly shifted for better illustration. All tiles exhibit overlap. During inference, tiles are merged halfway at their overlaps.
	}
	\label{figure:tiling}
\end{figure}

We now reframe our very general questions (a), (b) and (c) into two tractable research questions (1) and (2) that can be evaluated on our specific GPU hardware:

\textbf{(1) Fix Total Size, Vary Tile Size.}\footnote{Total size and tile size here refer to a number of pixels, not to a tuple $(width, height)$.}. Looking at a document image with a pixel count $N$, we ask ourselves, what is the difference between processing it as a whole vs. tiling it using h, v or hv patterns (see Figure \ref{figure:tiling})? In other words, what decrease in quality should we expect with tiled configurations of a fixed overall image resolution as opposed to processing the image as a whole?

\textbf{(2) Fix Tile Size, Vary Total Size.} Again, assuming some number of pixels $N$, we ask: what is the best use we can put these pixels to, i.e. should we train on a downscaled image of maximum size as a whole or should we use a bigger image (that does not fit on the GPU) and use tiles of maximum size?

While we could have evaluated (1) and (2) for many pixel counts $N$ and on both GPUs, because of the huge runtimes involved and to minimize GPU computation time, we chose to evaluate (1) only for $N=1.15M$ (i.e. the \emph{maximum size} of the V100 case), and (2) partially (namely for -, h, v) for $N=0.39M$ (i.e. the \emph{maximum size} of the GeForce RTX 1080 Ti case), and partially (namely for v and hv) for $N=1.15M$.

\begin{table*}\centering
\ra{1.1}
\begin{tabular}{@{}r|c|c|cc@{}}
configuration & total size & tile size & tiling \\
\toprule
\\

\multicolumn{1}{l}{\textbf{GeForce RTX 2080 Ti}}
\\

\midrule

\ldelim\{{3}{4cm}[\makecell{fixed\\\textbf{tile}\\size} ]

$0.3/-$ & $max$ & $max$ & - \\
% \textbf{0.39M} ($512 \times 768$)

$0.6/h$ & $> max$ & $max$ & h \\
% 0.66M ($640 \times 1024$)
% \textbf{0.39M} ($384 \times 1024$)

$0.9/v$ & $> max$ & $max$ & v \\
% 0.98M ($768 \times 1280$)
\\

\multicolumn{1}{l}{\textbf{V100}}

\\

\midrule

\ldelim\{{4}{4cm}[\makecell{fixed\\\textbf{total}\\size} ]
$1.1/-$ & $max$ & $max$ & - \\
$1.1/h$ & $max$ & $< max$ & h \\
$1.1/v$ & $max$ & $< max$ & v \\
$1.1/hv$ & $max$ & $< max$ & hv \\

\\
\midrule
\ldelim\{{2}{4cm}[\makecell{fixed \textbf{tile} size} ]
$3.0/v$ & $> max$ & $max$ & v \\
$3.9/hv$ & $> max$ & $max$ & hv \\

\bottomrule
\end{tabular}
\caption{\label{table:tilings-questions}Tiling configurations and related research questions, grouped by GPUs. $max$ indicates the maximum number of pixels for the respective GPU. For tiling patterns referenced here, see Figure \ref{figure:tiling}. As described in the text, configuration names consist of a prefix (e.g. $1.1$) indicating overall total size as pixel count followed by a tile pattern.}
\end{table*}

	\begin{table*}
	\centering
	\ra{1.1}
	\begin{tabular}{@{}lccccccccc@{}}\toprule

	name & \multicolumn{2}{c}{total size} & \phantom{abc} & \multicolumn{2}{c}{tiling} & \multicolumn{2}{c}{tile size} \\ 
	\cmidrule{2-3} \cmidrule{5-8} 
	& resolution & \#pixels & & pattern & \#tiles & resolution & \#pixels & \\
\midrule

$0.3/-$ & $512 \times 768$ & \textbf{0.39M} &  & $-$ & $-$ & $-$ & $-$\\
$0.6/h$ & $640 \times 1024$ & 0.66M &  & h & 2 & $384 \times 1024$ & \textbf{0.39M}\\
$0.9/v$ & $768 \times 1280$ & 0.98M &  & v & 3 & $768 \times 512$ & \textbf{0.39M}\\
&&&\\
$1.1/h$ & $896 \times 1280$ & \textbf{1.15M} & & h & 5 & $256 \times 1280$ & 0.33M \\
$1.1/v$ & $896 \times 1280$ & \textbf{1.15M} & & v & 4 & $896 \times 384$ & 0.25M \\
$1.1/hv$ & $896 \times 1280$ & \textbf{1.15M} &  & h, v & 4 & $512 \times 768$ & \textbf{0.39M}\\
$1.1/-$ & $896 \times 1280$ & \textbf{1.15M} &  & $-$ & $-$ & $-$ & $-$\\
&&&\\
$3.0/v$ & $1280 \times 2400$ & 3.07M &  & v & 3 & $1280 \times 896$ & 1.15M\\
$3.9/hv$ & $1640 \times 2400$ & 3.94M &  & h, v & 4 & $896 \times 1280$ & 1.15M\\

	\end{tabular}
\caption{\label{table:tilings}Details of the Tiling configurations.  \emph{Maximum sizes} are printed in bold.}
	\end{table*}

\FloatBarrier

Table \ref{table:tilings-questions} gives a concise overview of the main ideas just described. Table \ref{table:tilings} gives more detailed information on the resolutions used. In both tables we use a naming scheme for tile configurations that consists of a prefix indicating the overall size of the scaled, untiled document (e.g., by definition, $1.15$ for a total size of 1.15 megapixels) followed by the kind of tiling applied on it (e.g. $/h$ indicating some kind of horizontal tiling). This notation is clearly inexact, as it does not specify the tile size, and is therefore only intended as a shorthand for the more detailed data in Tables \ref{table:tilings-questions} and \ref{table:tilings}.

As illustration of the notation, three specific scenarios from question (2) are as follows: (2a) use all pixels in one image without tiling ($0.39/-$), (2b) use a bigger sized image and use all available pixels for horizontal tiles ($0.66/h$), (2c) use an even bigger sized image and use available pixels for vertical tiles ($0.98/v$).

\subsubsection{Results of Grid Search}

The results of our grid search over the described configuration space are given in Table \ref{table:gridresults}. Without preliminary studies and without dhSegment (which we evaluated separately), we evaluated $9 \times 10 \times 3 \times 2$ (each configuration, each backbone, three tasks, for both U-Net and FPN) $= 540$ data points. On average, each data point took about 15 hours of GPU time to compute.

\begin{table}
    \centering
    \scriptsize
	\ra{1.0}
	
	\begin{tabular}{@{}lccccccccc@{}}

	configuration & $0.3/-$ & $0.6/h$ & $0.9/v$ & $1.1/h$ & $1.1/v$ & $1.1/hv$ & $1.1/-$ & $3.0/v$ & $3.9/hv$ \\
	\toprule
	\\
	
results for \emph{sep}:\
\\
\midrule

			VGG16 & 91.29 & 91.67 & 91.51 & 82.74 & 90.77 & 90.58 & 90.59 & 90.87 & 89.52\\
ResNet-34 & 89.45 & 90.60 & 90.85 & 82.46 & 90.54 & 90.34 & 90.24 & 90.72 & 89.82\\
ResNet-50 & 89.93 & 90.94 & 91.43 & 82.61 & 90.22 & 90.41 & 90.45 & 90.79 & 89.63\\
SE-ResNet-34 & 89.91 & 91.17 & 91.12 & 82.57 & 90.32 & 90.57 & 90.18 & 91.19 & 89.82\\
SE-ResNet-50 & 90.73 & 91.32 & 91.70 & 82.95 & 90.34 & 90.75 & 90.69 & 91.44 & 89.92\\
SE-ResNeXt-50 & 90.77 & 91.45 & 91.62 & 82.97 & 90.88 & 90.81 & 90.03 & 91.24 & 90.05\\
EfficientNet-B1 & 91.36 & \textbf{92.01} & 91.84 & 83.17 & \textbf{91.09} & 91.08 & \textbf{91.18} & 91.64 & 90.21\\
EfficientNet-B2 & \textbf{91.63} & 91.70 & \textbf{91.86} & \textbf{83.34} & 91.04 & \textbf{91.15} & 91.07 & \textbf{91.70} & \textbf{90.27}\\
Inception-v3 & 91.06 & 91.42 & 91.19 & 82.80 & 90.69 & 90.96 & 89.99 & 91.21 & 89.84\\
Inception-ResNet-v2 & 91.18 & 91.64 & 91.41 & 82.92 & 90.70 & 90.85 & 90.94 & 91.09 & 89.96\\
dhSegment & 65.24 & 80.42 & 77.57 & 80.79 & 75.92 & 74.91 & 69.69 & 73.47 & 83.52\\

\\
results for \emph{blk}:\\
\midrule

			VGG16 & 92.01 & 93.06 & 93.81 & 91.21 & 94.07 & 93.84 & 93.24 & 94.68 & 93.82\\
ResNet-34 & 93.56 & 93.59 & 94.20 & 91.97 & 94.27 & 94.00 & 94.33 & 95.00 & 94.50\\
ResNet-50 & 93.21 & 93.39 & 94.58 & 91.58 & 94.60 & 94.06 & 94.79 & 95.04 & 94.45\\
SE-ResNet-34 & 92.49 & 93.50 & 93.85 & 91.49 & 94.39 & 94.22 & 94.10 & 95.16 & 94.42\\
SE-ResNet-50 & 93.53 & 93.85 & 94.59 & 91.82 & 94.79 & 94.60 & 94.39 & 94.86 & 94.61\\
SE-ResNeXt-50 & 93.61 & 93.88 & \textbf{94.93} & 92.19 & 94.69 & \textbf{94.65} & 94.73 & 95.05 & 94.61\\
EfficientNet-B1 & 92.83 & 93.38 & 94.17 & 91.63 & 94.23 & 94.05 & 93.74 & 94.96 & 94.56\\
EfficientNet-B2 & 92.59 & 93.07 & 94.11 & 91.81 & 94.18 & 94.07 & 94.19 & 95.03 & 94.59\\
Inception-v3 & 93.15 & 93.25 & 93.70 & 91.59 & 94.52 & 94.24 & 94.53 & 95.30 & 94.76\\
Inception-ResNet-v2 & \textbf{94.14} & 93.98 & 94.85 & 91.64 & \textbf{95.01} & 94.47 & \textbf{95.07} & \textbf{95.79} & \textbf{95.32}\\
dhSegment & 92.95 & \textbf{94.03} & 94.48 & \textbf{94.09} & 94.47 & 94.17 & 93.41 & 94.37 & 94.36\\

\\
results for \emph{blkx}:\\
\midrule

			VGG16 & 91.50 & 93.32 & 93.61 & 91.03 & 93.44 & 93.37 & 93.23 & 93.94 & 93.21\\
ResNet-34 & 92.70 & 93.09 & 93.57 & 91.15 & 94.11 & 93.33 & 93.54 & 94.65 & 93.73\\
ResNet-50 & 92.42 & 93.28 & 94.03 & 91.29 & 94.17 & 93.66 & 93.80 & 94.59 & 93.81\\
SE-ResNet-34 & 92.59 & 92.83 & 94.08 & 91.43 & 93.93 & 93.62 & 93.68 & 94.60 & 94.41\\
SE-ResNet-50 & 92.60 & 93.70 & 94.11 & 91.65 & 94.30 & 93.90 & 94.27 & 94.67 & 94.14\\
SE-ResNeXt-50 & \textbf{93.37} & 93.80 & 94.19 & 91.59 & 94.16 & 94.36 & 94.29 & 94.52 & 93.95\\
EfficientNet-B1 & 92.32 & 92.96 & 94.02 & 90.90 & 93.65 & 93.77 & 93.60 & 94.61 & 94.28\\
EfficientNet-B2 & 92.26 & 93.11 & 93.65 & 91.35 & 93.82 & 93.60 & 93.87 & 94.85 & 94.27\\
Inception-v3 & 92.99 & 93.38 & 94.01 & 91.50 & 94.02 & 93.88 & 93.99 & 95.08 & 94.34\\
Inception-ResNet-v2 & 93.32 & 93.73 & \textbf{94.78} & 91.71 & \textbf{94.65} & \textbf{94.58} & \textbf{94.60} & \textbf{95.51} & \textbf{94.91}\\
dhSegment & 92.91 & \textbf{94.19} & 94.44 & \textbf{94.04} & 94.29 & 94.03 & 93.26 & 94.35 & 94.25\\

\bottomrule
\end{tabular}
\caption{Matthew correlation coefficients (MCCs) all evaluated models in grid search. Column titles are the tile configuration names described in Table \ref{table:tilings}. Shown results include $\Theta_{s-model}$ with both U-Net and FPN (only best variant is shown). Best values in each column are bold (per task).}
\label{table:gridresults}
\end{table}

The actual numbers make a simple comparison difficult, since they are relative to resolution. Seeing a similar number between a smaller resolution or tile size and a larger resolution does not indicate that the result on the larger resolution is worse, it might in fact be even better. On the other hand, a higher number on a higher resolution configuration (i.e. a more difficult problem) should always indicate a better result.

\FloatBarrier

\begin{figure}[h]

    \centering 
    % This file was created by tikzplotlib v0.9.1.
\begin{tikzpicture}[scale=0.75]

\definecolor{color0}{rgb}{0.196078431372549,0.803921568627451,0.196078431372549}
\definecolor{color1}{rgb}{0.67843137254902,0.847058823529412,0.901960784313726}
\definecolor{color2}{rgb}{0.580392156862745,0,0.827450980392157}
\definecolor{color3}{rgb}{0.803921568627451,0.52156862745098,0.247058823529412}
\definecolor{color4}{rgb}{0.12156862745098,0.466666666666667,0.705882352941177}
\definecolor{color5}{rgb}{1,0.498039215686275,0.0549019607843137}
\definecolor{color6}{rgb}{0.172549019607843,0.627450980392157,0.172549019607843}

\begin{groupplot}[group style={group size=2 by 1}]
\nextgroupplot[
tick align=outside,
tick pos=left,
x grid style={white!69.0196078431373!black},
xmin=-0.465, xmax=9.765,
xtick style={color=black},
xtick={0,1,2,3,4,5,6,7,8,9},
xticklabel style = {rotate=90.0},
xticklabels={Inception-v3,SE-ResNet-34,Inception-ResNet-v2,EfficientNet-B2,VGG16,EfficientNet-B1,SE-ResNet-50,SE-ResNeXt-50,ResNet-50,ResNet-34},
y grid style={white!69.0196078431373!black},
ylabel={Quality (MCC)},
ymin=91.2376111111111, ymax=96.8023888888889,
ytick style={color=black}
]
\path [draw=color0, line width=0.82pt]
(axis cs:0,92.0383333333333)
--(axis cs:0,95.7483333333333);

\path [draw=color1, line width=0.82pt]
(axis cs:1,91.9005555555555)
--(axis cs:1,95.5705555555556);

\path [draw=color0, line width=0.82pt]
(axis cs:2,92.3994444444444)
--(axis cs:2,96.5494444444444);

\path [draw=color2, line width=0.82pt]
(axis cs:3,92.1277777777778)
--(axis cs:3,95.3477777777778);

\path [draw=color3, line width=0.82pt]
(axis cs:4,91.5694444444444)
--(axis cs:4,95.0394444444445);

\path [draw=color2, line width=0.82pt]
(axis cs:5,92.0627777777778)
--(axis cs:5,95.3927777777778);

\path [draw=color1, line width=0.82pt]
(axis cs:6,92.5955555555555)
--(axis cs:6,95.6355555555556);

\path [draw=color1, line width=0.82pt]
(axis cs:7,92.83)
--(axis cs:7,95.69);

\path [draw=color1, line width=0.82pt]
(axis cs:8,92.2366666666667)
--(axis cs:8,95.6966666666667);

\path [draw=color1, line width=0.82pt]
(axis cs:9,92.4205555555556)
--(axis cs:9,95.4505555555556);

\path [draw=color0, line width=1.64pt]
(axis cs:0.3,91.8977777777778)
--(axis cs:0.3,95.4777777777778);

\path [draw=color1, line width=1.64pt]
(axis cs:1.3,91.8783333333333)
--(axis cs:1.3,95.0483333333333);

\path [draw=color0, line width=1.64pt]
(axis cs:2.3,92.2988888888889)
--(axis cs:2.3,96.0988888888889);

\path [draw=color2, line width=1.64pt]
(axis cs:3.3,91.67)
--(axis cs:3.3,95.17);

\path [draw=color3, line width=1.64pt]
(axis cs:4.3,91.5061111111111)
--(axis cs:4.3,94.4161111111111);

\path [draw=color2, line width=1.64pt]
(axis cs:5.3,91.4905555555555)
--(axis cs:5.3,95.2005555555556);

\path [draw=color1, line width=1.64pt]
(axis cs:6.3,92.1944444444445)
--(axis cs:6.3,95.2144444444444);

\path [draw=color1, line width=1.64pt]
(axis cs:7.3,92.3383333333333)
--(axis cs:7.3,95.2683333333333);

\path [draw=color1, line width=1.64pt]
(axis cs:8.3,91.8)
--(axis cs:8.3,95.1);

\path [draw=color1, line width=1.64pt]
(axis cs:9.3,91.5688888888889)
--(axis cs:9.3,95.0688888888889);

\addplot [line width=0.82pt, black, mark=+, mark size=3, mark options={solid}, only marks]
table {%
0 93.8933333333333
1 93.7355555555555
2 94.4744444444444
3 93.7377777777778
4 93.3044444444444
5 93.7277777777778
6 94.1155555555555
7 94.26
8 93.9666666666667
9 93.9355555555556
};
\addplot [line width=1.64pt, black, mark=+, mark size=3, mark options={solid}, only marks]
table {%
0.3 93.6877777777778
1.3 93.4633333333333
2.3 94.1988888888889
3.3 93.42
4.3 92.9611111111111
5.3 93.3455555555555
6.3 93.7044444444444
7.3 93.8033333333333
8.3 93.45
9.3 93.3188888888889
};

\nextgroupplot[
legend cell align={left},
legend style={fill opacity=0.8, draw opacity=1, text opacity=1, at={(0.97,0.03)}, anchor=south east, draw=white!80!black},
tick align=outside,
tick pos=left,
x grid style={white!69.0196078431373!black},
xlabel={Number of training pages},
xmin=4.25, xmax=86.75,
xtick style={color=black},
y grid style={white!69.0196078431373!black},
ylabel={},
ymin=87.8874131308627, ymax=96.0925532630593,
ytick style={color=black}
]
\addplot [line width=1.64pt, color4, mark=+, mark size=3, mark options={solid}]
table {%
8 88.2603740459625
16 90.0857560261357
24 90.4732402904983
33 90.6799742465348
41 91.3849750963741
49 91.2370043120032
58 91.1056447464935
66 91.478420856269
74 91.5537199634443
83 91.7500495513392
};
\addlegendentry{sep}
\addplot [line width=1.64pt, color5, mark=+, mark size=3, mark options={solid}]
table {%
8 91.0243447758816
16 93.3855802468238
24 94.5655961217516
33 94.6841520256142
41 94.8500367603432
49 94.9423789710124
58 95.2098663992341
66 95.3333380913203
74 95.399481974426
83 95.4342809983583
};
\addlegendentry{blkx}
\addplot [line width=1.64pt, color6, mark=+, mark size=3, mark options={solid}]
table {%
8 91.3683587208783
16 93.6481075608274
24 94.5618218864974
33 94.5944443987269
41 95.3017280802972
49 95.1348937186423
58 95.4122989202557
66 95.7195923479595
74 95.6787356036584
83 95.5523077944613
};
\addlegendentry{blk}
\end{groupplot}

\end{tikzpicture}

	\caption{Left: Minimum, maximum and mean MCC (over all tiling configurations) for several backbones (color coded by family, e.g. Inception is green) regarding \emph{blk} (thin lines, left) and \emph{blkx} (thick lines, right). Sorted from lowest absolute mean decrease when going from \emph{blk} to \emph{blkx} (left) to largest decrease (right).
	Right: Quality of trained models vs. number of training pages used (using a constant test set). This analysis was done on configuration $3.0/v$ with Inception-ResNet-v2.
	}
	\label{figure:figure-ablation}
\end{figure}
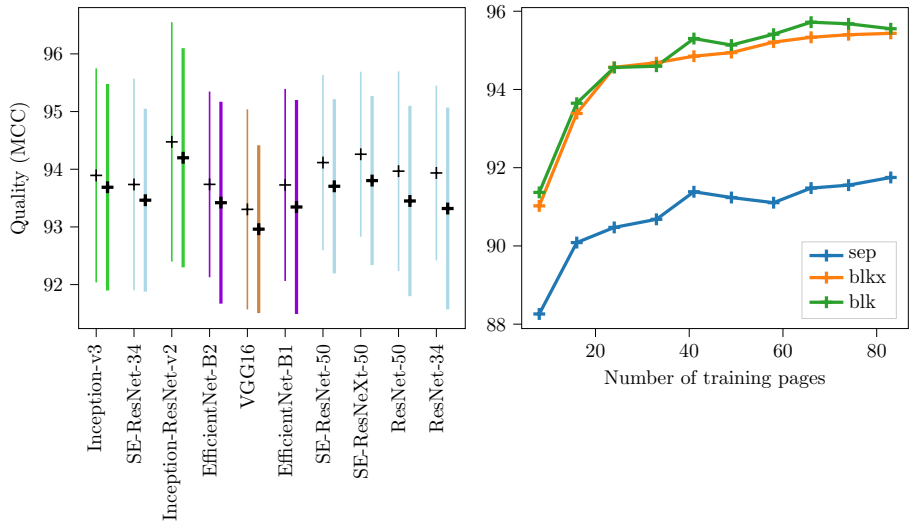

With these caveats in mind, here are some observations on this data:
\vspace{0.25cm}

\textbf{Tasks.} In general, \emph{blk} yields better results (higher scores) than \emph{blkx}, which is expected, as \emph{blk} is a simpler task. See Figure \ref{figure:figure-ablation} for how much better the results are for \emph{blk}.

\vspace{0.2cm}
\textbf{Backbone Architecture.} For the \emph{sep} task, EfficientNet outperforms all other architectures.\footnote{All models, expect for dhSegment, used dice loss for this task. This might be a factor in this result.}.

For \emph{blk}, \emph{blkx} we see three notable architectures: SE-ResNeXt-50, Inception-ResNet-v2, and dhSegment. The latter performs better (but not great) on h tilings than the other architectures, which might itself be an artefact due to dhSegment's internal scaling\footnote{dhSegment seems to process images at a higher resolution internally than they are given.}. In general, Inception-ResNet-v2 (which is in fact the best backbone architecture among the tested ones \cite{bianco_benchmark_2018}) seems to outperform all other architectures as the complexity of task gets harder (going from \emph{blk} to \emph{blkx}), or as the resolution of the image gets higher. We also observe that more powerful backbone architectures seem to expose a less severe drop of performance when going from \emph{blk} to \emph{blkx} (see Figure \ref{figure:figure-ablation}).

\textbf{S-Model Architectures.} We investigated both FPN and U-Net, but did not find systematic or large performance differences between the two. That said, our final models (based on $3.0/v$) achieved their best scores on all three tasks with FPN.

\begin{figure}[h]
    \centering 
    
    \centering
    % This file was created by tikzplotlib v0.9.1.
\begin{tikzpicture}[scale=0.75]

\definecolor{color0}{rgb}{0.12156862745098,0.466666666666667,0.705882352941177}
\definecolor{color1}{rgb}{1,0.498039215686275,0.0549019607843137}
\definecolor{color2}{rgb}{0.737254901960784,0.741176470588235,0.133333333333333}
\definecolor{color3}{rgb}{0.549019607843137,0.337254901960784,0.294117647058824}

\begin{groupplot}[group style={group size=2 by 1}]
\nextgroupplot[
tick align=outside,
tick pos=left,
x grid style={white!69.0196078431373!black},
xlabel={Image resolution (megapixels)},
xmin=0.2125, xmax=4.1175,
xtick style={color=black},
y grid style={white!69.0196078431373!black},
ylabel={Quality (MCC)},
ymin=90.6695, ymax=95.7405,
ytick style={color=black}
]
\path [fill=color0, fill opacity=0.3]
(axis cs:0.66,92.83)
--(axis cs:0.66,94.19)
--(axis cs:1.15,94.04)
--(axis cs:1.15,90.9)
--cycle;
\path [fill=color1, fill opacity=0.3]
(axis cs:0.98,93.57)
--(axis cs:0.98,94.78)
--(axis cs:1.15,94.65)
--(axis cs:1.15,93.65)
--cycle;
\path [fill=color1, fill opacity=0.3]
(axis cs:1.15,93.65)
--(axis cs:1.15,94.65)
--(axis cs:3.07,95.51)
--(axis cs:3.07,94.35)
--cycle;
\path [fill=color2, fill opacity=0.3]
(axis cs:1.15,93.33)
--(axis cs:1.15,94.58)
--(axis cs:3.94,94.91)
--(axis cs:3.94,93.73)
--cycle;
\path [fill=color3, fill opacity=0.3]
(axis cs:0.39,92.26)
--(axis cs:0.39,93.37)
--(axis cs:1.15,94.6)
--(axis cs:1.15,93.26)
--cycle;

\addplot [line width=1.64pt, color0]
table {%
0.66 94.19
1.15 94.04
};
\addplot [line width=1.64pt, color0, dotted]
table {%
0.66 92.83
1.15 90.9
};
\addplot [line width=1.64pt, color1]
table {%
0.98 94.78
1.15 94.65
};
\addplot [line width=1.64pt, color1, dotted]
table {%
0.98 93.57
1.15 93.65
};
\addplot [line width=1.64pt, color1]
table {%
1.15 94.65
3.07 95.51
};
\addplot [line width=1.64pt, color1, dotted]
table {%
1.15 93.65
3.07 94.35
};
\addplot [line width=1.64pt, color2]
table {%
1.15 94.58
3.94 94.91
};
\addplot [line width=1.64pt, color2, dotted]
table {%
1.15 93.33
3.94 93.73
};
\addplot [line width=1.64pt, color3]
table {%
0.39 93.37
1.15 94.6
};
\addplot [line width=1.64pt, color3, dotted]
table {%
0.39 92.26
1.15 93.26
};

\nextgroupplot[
legend cell align={left},
legend style={fill opacity=0.8, draw opacity=1, text opacity=1, at={(0.97,0.03)}, anchor=south east, draw=white!80!black},
tick align=outside,
tick pos=left,
x grid style={white!69.0196078431373!black},
xlabel={Image resolution (megapixels)},
xmin=0.2125, xmax=4.1175,
xtick style={color=black},
y grid style={white!69.0196078431373!black},
ymin=82.2935, ymax=92.1165,
ytick style={color=black}
]
\addplot [line width=0.82pt, color0, forget plot]
table {%
0.66 91.67
1.15 82.74
};
\addplot [line width=0.82pt, color1, forget plot]
table {%
0.98 91.51
1.15 90.77
3.07 90.87
};
\addplot [line width=0.82pt, color2, forget plot]
table {%
1.15 90.58
3.94 89.52
};
\addplot [line width=0.82pt, color3, forget plot]
table {%
0.39 91.29
1.15 90.59
};
\end{groupplot}

\matrix[
    matrix of nodes,
    anchor=north west,
    inner sep=0.3em,
    column 1/.style={nodes={anchor=east}},
    column 2/.style={nodes={anchor=west},font=\strut}
  ]
  at([xshift=-2.5cm, yshift=-2cm]current axis.north east){
  	\node[rectangle, fill=color0, text=white, draw=black] { }; & h\\
  	\node[rectangle, fill=color1, text=white, draw=black] { }; & v\\
  	\node[rectangle, fill=color2, text=white, draw=black] { }; & hv\\
  	\node[rectangle, fill=color3, text=white, draw=black] { }; & untiled\\
    };

\end{tikzpicture}

    \caption{Left: Solid areas of best (solid lines) and worst (dotted lines) performance of all \emph{blkx} models in our grid search (without VGG16). Except for h, performance increases as overall image resolution increases, most notably for untiled and for v. Right: Performance of backbone VGG16, which bucks the general trend and tends to get worse as image resolution increases.
	}
    \label{figure:resolutionv}

\end{figure}
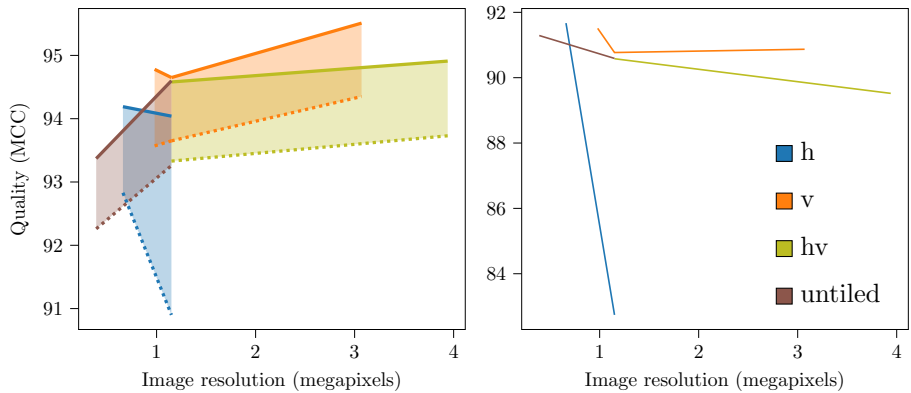

\vspace{0.2cm}
\textbf{Tiling and Resolution.} We find that higher resolutions generally seems to improve results, but observe a strong dependency on the mode of tiling used on how big that gain actually is, with v tiling showing the highest relative increase per resolution increase (see left plot in Figure \ref{figure:resolutionv}). Notably, this general observation does not hold true for h tiling and for VGG16. The latter is the only tested model that decreases performance for the hv and even the untiled mode with increased resolution and shows no clear increase for v (see right plot in Figure \ref{figure:resolutionv}).

Our research questions concerning tiling can be answered as follows:
\vspace{0.25cm}

(1) when fixing total size while varying tile size, we see different results for different tasks. For \emph{sep} and \emph{blkx}, tile patterns v, hv and "no tiling" are close, whereas h tiling is clearly worse (see $1.1$, $1.1/h$, $1.1/v$, $1.1/hv$). For \emph{blk}, on the other hand, only v-tiling is close to the untiled version.

(2) when fixing tile size while varying total size, vertical tiles seem to work best for \emph{blk} and \emph{blkx} (see $0.9/v$ vs. $0.3/-$ and $0.6/h$). For \emph{sep}, on the other hand, this does not seem true, as indeed $0.6/h$ performs best for various backbones.\footnote{$0.6/h$ uses 2 h-tiles, whereas $1.1/h$ uses 5 h-tiles (see Table \ref{table:tilings}). We assume that this gives $0.6/h$ very good coverage of long vertical separator lines as well as a reasonable coverage of horizontal separator lines, whereas $1.1/h$ gets bad coverage of the latter.}

Still, taking into account all tasks and the bad results for $1.1/h$, we argue that v-tiling seems to be the best general option for training newspaper data sets, as it yields results that are consistently close to untiled training across various tasks.

\vspace{0.2cm}
\textbf{Partial Data.} Additional experiments with using only parts of our training data indicate that a training data set of 20 to 40 pages already gives very good results (see Figure \ref{figure:figure-ablation}). More specifically, we already see $99\%$ of the performance of our full (84 page) training set at 24 pages for \emph{blkx} and at 41 pages for \emph{sep}. Thus, for \emph{blkx}, tripling the amount of training data from 24 pages to 72 pages only yields a relative gain in MCC of less than $1\%$.

\FloatBarrier

\subsection{Final Model}

In Table \ref{table:evaluation}, we show results for our final models. We also reprint the results from \cite{chen_convolutional_2017} to illustrate that our approach seems to work reasonably well in its domain. Note that \cite{wick_fully_2018} also report a pixel accuracy of 98.4\% on St. Gall. Inference for one fold takes about $0.63 s$ per page on the V100. Following our reliance on MCC, we pick \emph{sep}/fold2 (and not \emph{sep}/fold4) as best fold for \emph{sep}.
 
\begin{table}[h]
\begin{center}
  \begin{tabular}{ r l c c c c c }
	task & fold & pixel accuracy & mean acc. & mean IU & f.w. IU & MCC \\
    \hline

\emph{sep}&1 & 99.76 & 87.06 & 81.24 & 99.58 & 91.51\\
\emph{sep}&2 & 99.77 & \textbf{89.58} & \textbf{85.03} & 99.58 & \textbf{92.08}\\
\emph{sep}&3 & 99.72 & 85.83 & 79.61 & 99.52 & 89.85\\
\emph{sep}&4 & \textbf{99.79} & 86.90 & 81.50 & \textbf{99.62} & 91.39\\
\emph{sep}&5 & 99.73 & 89.48 & 83.49 & 99.51 & 90.24\\

    &&&&&\\

\emph{blkx}&1 & 97.58 & 77.16 & 74.86 & 95.52 & 95.76\\
\emph{blkx}&2 & 97.96 & 88.77 & 85.88 & 96.10 & 96.14\\
\emph{blkx}&3 & 97.68 & 89.22 & 86.30 & 95.63 & 95.86\\
\emph{blkx}&4 & \textbf{98.86} & \textbf{94.53} & \textbf{92.67} & \textbf{97.82} & \textbf{97.83}\\
\emph{blkx}&5 & 98.17 & 88.55 & 85.86 & 96.78 & 96.35\\

    &&&&&\\
    \hline
	
	\multicolumn{2}{r}{G. Washington \cite{chen_convolutional_2017}} & 91 & 91 & 77 & 86 & - \\
	\multicolumn{2}{r}{Parzival \cite{chen_convolutional_2017}} & 94 & 75 & 68 & 89 & - \\
	\multicolumn{2}{r}{St. Gall \cite{chen_convolutional_2017}} & 98 & 90 & 87 & 96 & - \\
	
  \end{tabular}
\end{center}
\caption{\label{table:evaluation}Final $3.0/v$ models and reference scores from \cite{chen_convolutional_2017}. Best scores in each columns (and for each task) are bold.}
\end{table}

\FloatBarrier

\newpage
\section{Conclusion}

We were able to re-support the claim that pixel-by-pixel segmentation approaches can be applied very successfully to newspapers, and furthermore showed that transfer learning works well in this context. We also showed that detection of table areas and separators works reasonably well for even complex layouts.
Furthermore, we investigated the influence of various factors, such as tiling configurations or the influence of number of labels (i.e. \emph{blk} vs \emph{blkx}) and found data supporting the idea that vertical tilings work well.
Finally, we showed that a training data set of 30 to 40 pages already gives reasonably good results.

Our study is intended to be a starting point for further research with more carefully chosen tile and resolution subsets. We hope our data can offer valuable guidelines for groups working on similar problems.

\newpage

\bibliographystyle{unsrt}
\bibliography{main}

\newpage
\section*{Appendix}
\noindent

\begin{minipage}{\textwidth}
    \centering
    \begin{minipage}[b]{0.32\textwidth}
        \includegraphics[width=\textwidth]{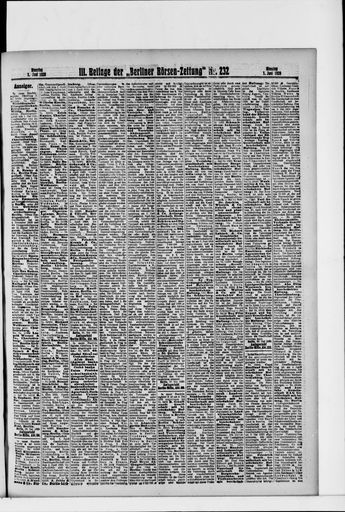}
    \end{minipage}
    ~ %add desired spacing between images, e. g. ~, \quad, \qquad, \hfill etc. 
      %(or a blank line to force the subfigure onto a new line)
    \begin{minipage}[b]{0.25\textwidth}
        \includegraphics[width=\textwidth]{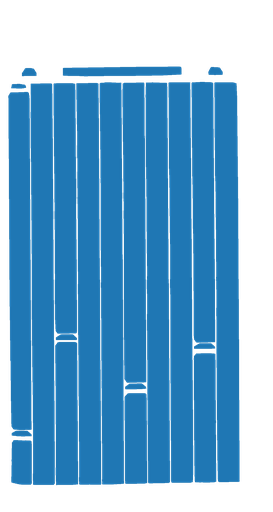}
    \end{minipage}
    ~ %add desired spacing between images, e. g. ~, \quad, \qquad, \hfill etc. 
    %(or a blank line to force the subfigure onto a new line)
    \begin{minipage}[b]{0.25\textwidth}
        \includegraphics[width=\textwidth]{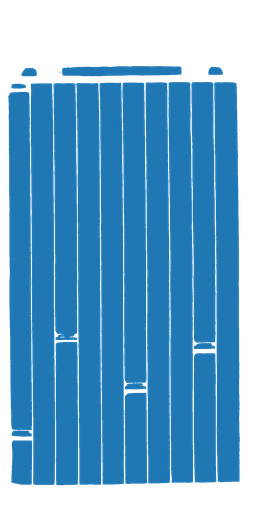}
    \end{minipage}
    
    \centering
    \begin{minipage}[b]{0.32\textwidth}
        \includegraphics[width=\textwidth]{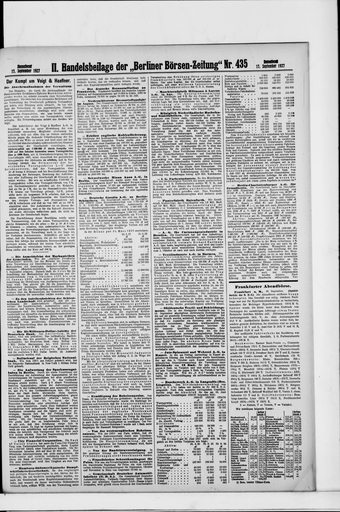}
    \end{minipage}
    ~ %add desired spacing between images, e. g. ~, \quad, \qquad, \hfill etc. 
      %(or a blank line to force the subfigure onto a new line)
    \begin{minipage}[b]{0.25\textwidth}
        \includegraphics[width=\textwidth]{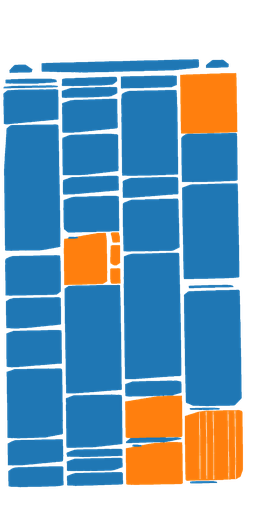}
    \end{minipage}
    ~ %add desired spacing between images, e. g. ~, \quad, \qquad, \hfill etc. 
    %(or a blank line to force the subfigure onto a new line)
    \begin{minipage}[b]{0.25\textwidth}
        \includegraphics[width=\textwidth]{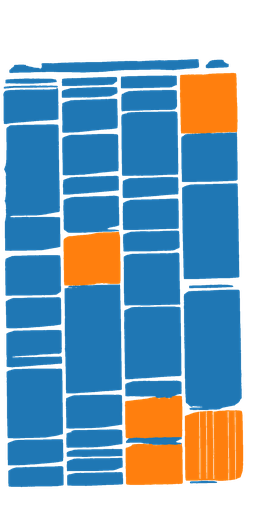}
    \end{minipage}

    \centering
    \begin{minipage}[b]{0.32\textwidth}
        \includegraphics[width=\textwidth]{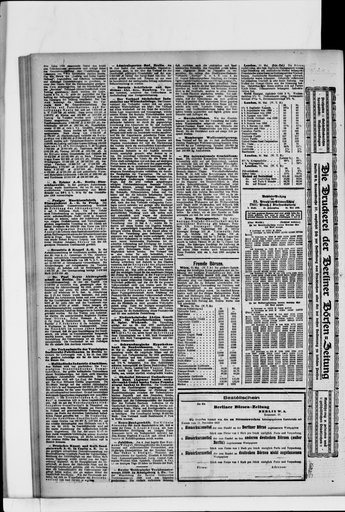}
    \end{minipage}
    ~ %add desired spacing between images, e. g. ~, \quad, \qquad, \hfill etc. 
      %(or a blank line to force the subfigure onto a new line)
    \begin{minipage}[b]{0.25\textwidth}
        \includegraphics[width=\textwidth]{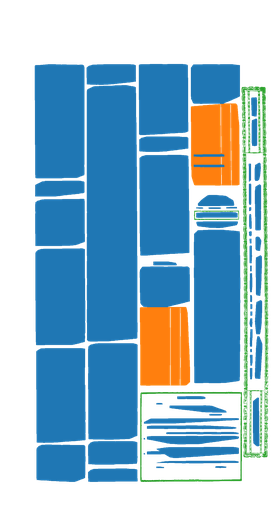}
    \end{minipage}
    ~ %add desired spacing between images, e. g. ~, \quad, \qquad, \hfill etc. 
    %(or a blank line to force the subfigure onto a new line)
    \begin{minipage}[b]{0.25\textwidth}
        \includegraphics[width=\textwidth]{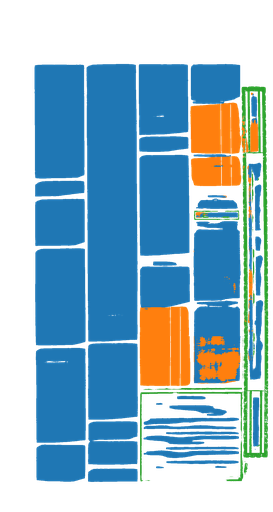}
    \end{minipage}
    
    \captionof{figure}{Examples from \emph{blkx}/fold-4. Left: Scanned page. Middle: Ground truth. Right: Prediction.
}\label{fig:results-blkx}

\end{minipage}

\begin{minipage}{\textwidth}
    \centering
    \begin{minipage}[b]{0.32\textwidth}
        \includegraphics[width=\textwidth]{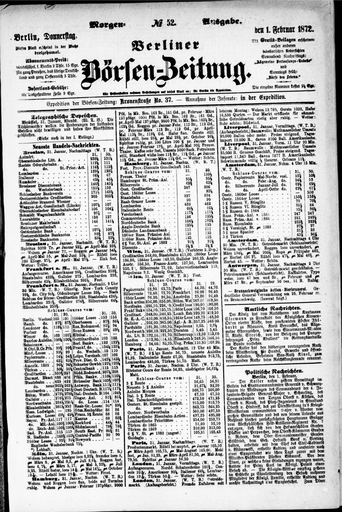}
    \end{minipage}
    ~ %add desired spacing between images, e. g. ~, \quad, \qquad, \hfill etc. 
      %(or a blank line to force the subfigure onto a new line)
    \begin{minipage}[b]{0.26\textwidth}
        \includegraphics[width=\textwidth]{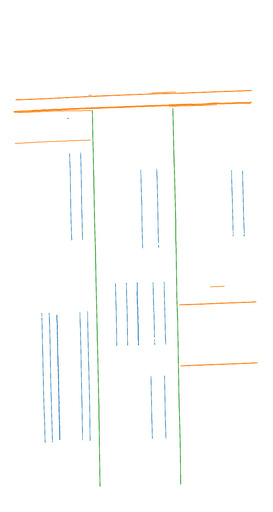}
    \end{minipage}
    \begin{minipage}[b]{0.26\textwidth}
        \includegraphics[width=\textwidth]{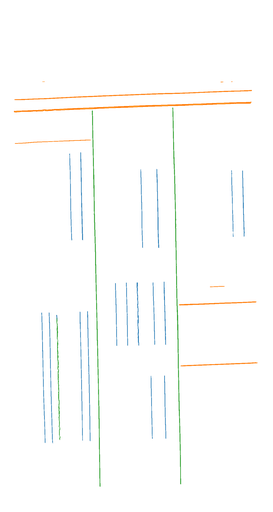}
    \end{minipage}
    
    \centering
    \begin{minipage}[b]{0.32\textwidth}
        \includegraphics[width=\textwidth]{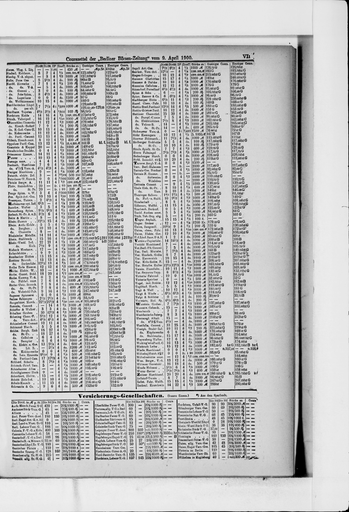}
    \end{minipage}
    \begin{minipage}[b]{0.26\textwidth}
        \includegraphics[width=\textwidth]{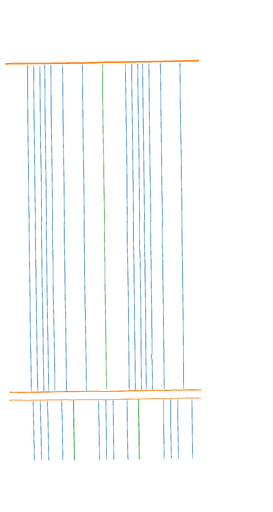}
    \end{minipage}
    \begin{minipage}[b]{0.26\textwidth}
        \includegraphics[width=\textwidth]{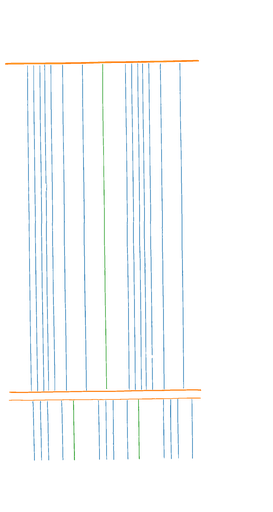}
    \end{minipage}

    \centering
    \begin{minipage}[b]{0.32\textwidth}
        \includegraphics[width=\textwidth]{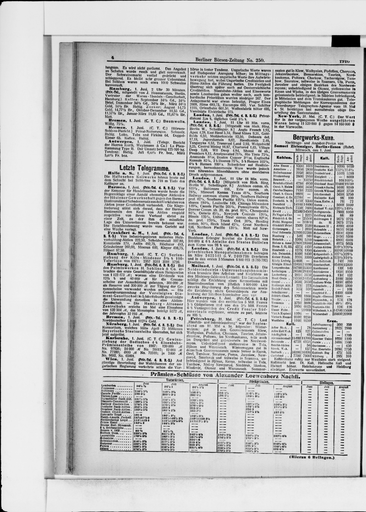}
    \end{minipage}
    \begin{minipage}[b]{0.26\textwidth}
        \includegraphics[width=\textwidth]{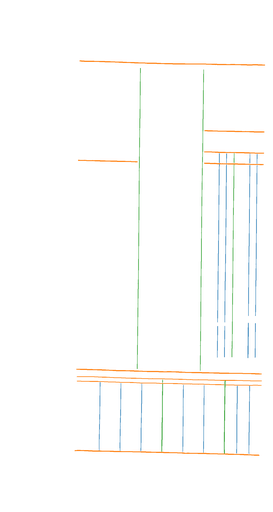}
    \end{minipage}
    \begin{minipage}[b]{0.26\textwidth}
        \includegraphics[width=\textwidth]{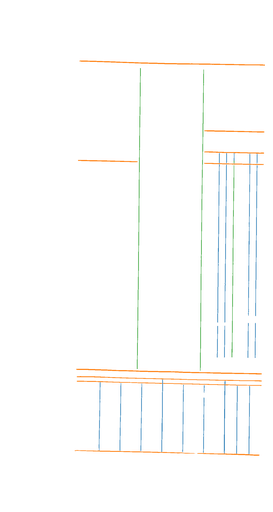}
    \end{minipage}
    
    \captionof{figure}{Examples from \emph{sep}/fold-2. Text separators show as green lines, table column separators as blue lines. Left: Scanned page. Middle: Ground truth. Right: Prediction.}
    \label{fig:results-sep}
\end{minipage}

\end{document}